\documentclass[conference]{IEEEtran}
\usepackage{cite,graphicx,amsmath,amssymb,amsfonts} 
\usepackage{graphicx}
\usepackage{amsmath,amssymb}
\usepackage{algorithm}
\usepackage{algpseudocode}
\usepackage{subcaption}
\usepackage{balance}
\usepackage{cite}
\usepackage{epstopdf}
\usepackage{multirow}
\title{Vision-Based Autonomous MM-Wave Reflector Using ArUco-Driven Angle-of-Arrival Estimation}

\author{
    Josue Marroquin, Nan Inzali, Miles Dillon Lantz,  Campbell Freeman,  Amod Ashtekar, \\Ajinkya Umesh Mulik, and Mohammed E Eltayeb \\
        Department of Electrical and Electronic Engineering, \\
    California State University, Sacramento \\
    Emails: \{josuemarroquin, ninzali, mlantz, cfreeman3, amodashtekar, aumulik, mohammed.eltayeb\}@csus.edu
}

\begin{document}

\maketitle

\begin{abstract}
Reliable millimeter-wave (mmWave) communication in non-line-of-sight (NLoS) conditions remains a key challenge for next-generation wireless systems, especially in indoor and infrastructure-limited settings. This paper presents a proof-of-concept vision-guided passive reflector system that enhances mmWave links by autonomously steering signal reflections with a motorized metallic plate. The system uses a lightweight monocular camera to detect fiducial ArUco markers on transmitter and receiver nodes, estimate their relative orientation, and align the reflector in real time without the need for RF feedback, external infrastructure, or Reconfigurable Intelligent Surfaces (RIS). In addition to reflector steering, the use of uniquely identifiable markers enables selective alignment toward authorized users, addressing the user identification problem not considered in prior vision-aided RIS work. A low-cost prototype, implemented on a Raspberry Pi~4, demonstrates operation in dynamic indoor NLoS scenarios at 60\,GHz. Experiments show received power improvements of up to 17\,dB (with an average gain of $\approx$10\,dB) compared to the no-reflector baseline. These results highlight vision-guided reflectors as a practical and energy-efficient alternative to relays or RIS panels, with strong potential for short-range applications such as IoT, AR/VR, and consumer wireless connectivity.
\end{abstract}

\begin{IEEEkeywords}
Millimeter-Wave Communications, Passive Reflector Steering, Vision-Aided Systems,   Autonomous Platforms.
\end{IEEEkeywords}

\section{Introduction}
\label{sec:intro}

Millimeter-wave (mmWave) and terahertz (THz) bands offer unprecedented data rates and highly directional communication, making them attractive for next-generation wireless systems, including indoor Wi-Fi, 5G/6G, and immersive AR/VR applications \cite{jiang2024terahertz,harvey2019mmwave,kim2022joint,vaccari2024tracking}. However, practical deployment remains constrained by their reliance on line-of-sight (LoS) links and strong susceptibility to blockage from walls, furniture, and moving objects. These limitations motivate lightweight, adaptive solutions that can extend coverage in non-line-of-sight (NLoS) scenarios without heavy infrastructure or complex RF feedback.  

One widely studied approach is the use of \textit{Reconfigurable Intelligent Surfaces} (RIS), which enhance propagation by electronically controlling reflected waves \cite{liu2021reconfigurable,elmossallamy2020reconfigurable,ouyang2023cvris,jiang2023camera,pei2021ris}. While promising, RIS platforms require dense arrays of tunable elements, frequent channel estimation, and beam training, resulting in significant hardware complexity and power consumption. These requirements become prohibitive in dense or mobile networks.  

\begin{figure}[t]
    \centering
    \includegraphics[width=0.9\linewidth]{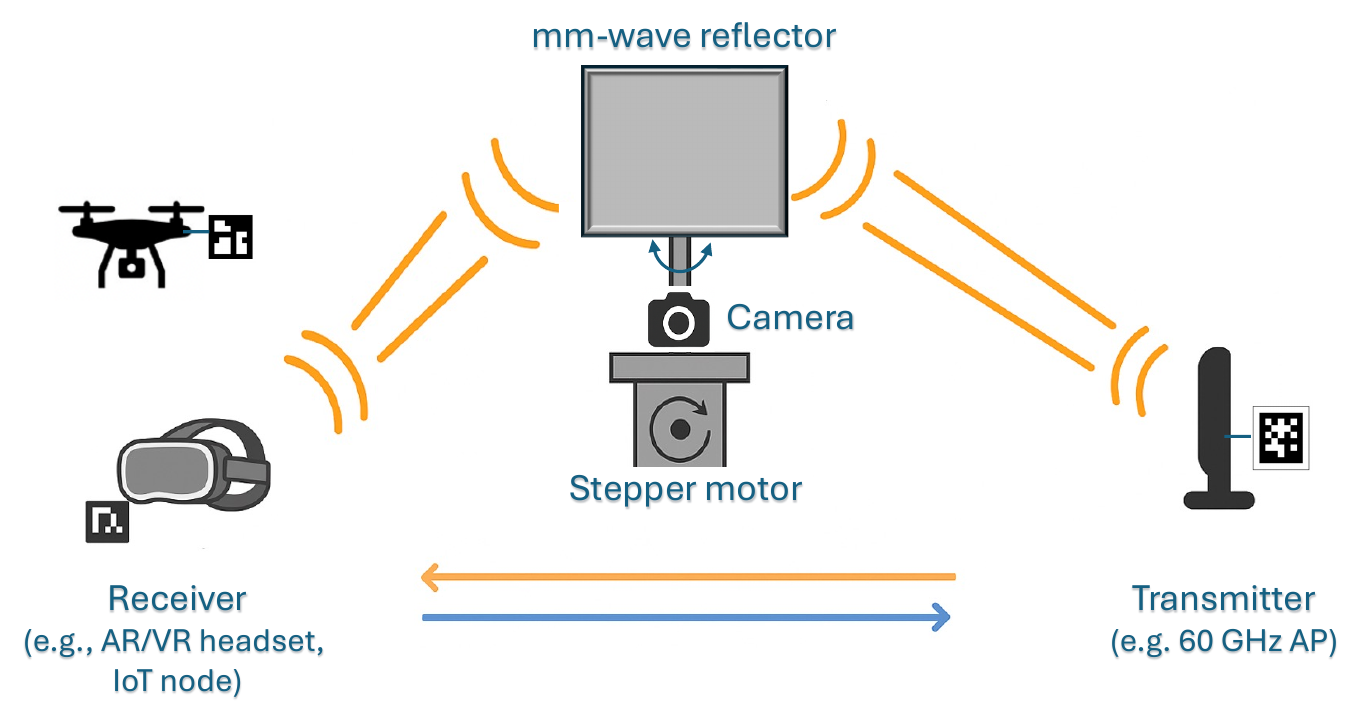}
    \caption{Conceptual illustration of the proposed vision-guided mmWave reflector system. 
    A motorized metallic reflector with an onboard camera dynamically redirects signals from a transmitter 
    (e.g., 60\,GHz access point) toward a receiver (e.g., AR/VR headset or IoT device) using fiducial markers for 
    real-time alignment.}
    \label{fig:systemmodel}
\end{figure}

Recently, \textit{vision-aided systems} have emerged as an alternative for beam management \cite{ouyang2023cvris,jiang2023camera,charan2021blockage,salehi2020machine}. By exploiting RGB or depth imagery to infer user position and orientation, they reduce or even eliminate the need for continuous RF feedback. For example, Ouyang et al.~\cite{ouyang2023cvris} integrated stereo cameras with a RIS panel for real-time beam steering; however, their framework assumes that the user’s initial location is perfectly known a priori before tracking begins. Other works, such as Jiang et al.~\cite{jiang2023camera}, applied monocular object detection to predict beam indices, further underscoring the potential of computer vision as a low-cost, scalable alternative to explicit RF probing.   In parallel, several studies have examined \textit{passive reflectors} as a simple, low-cost means of improving mmWave coverage in NLoS environments \cite{khawaja2020coverage,khawaja2019effect,ibrahim2025lidar}. While flat metallic reflectors can effectively redirect energy, their static orientation prevents adaptation to user mobility or environmental changes, limiting their practicality.  

To bridge the gap between static reflectors and complex RIS panels, we propose a \textit{vision-guided autonomous reflector system} that combines the infrastructure-free benefits of passive reflectors with the adaptability of computer vision. As shown in Fig.~\ref{fig:systemmodel}, the prototype consists of a flat metallic plate mounted on a two-axis gimbal steered by stepper motors and controlled by a Raspberry Pi~4. An onboard camera detects fiducial ArUco markers placed on transmitter and receiver nodes, computes their relative orientation, and dynamically adjusts the reflector to maximize specular redirection. Unlike RIS platforms, which require dense hardware arrays and continuous training, our approach operates without RF feedback, thereby reducing complexity, power consumption, and overhead.  Moreover, the use of ArUco markers not only guides reflector alignment but also enables user identification, allowing selective redirection toward authorized devices, an aspect not addressed in prior vision-aided RIS studies.

\textbf{\textit{Why ArUco?}} Unlike QR codes or generic object detectors, ArUco markers are explicitly designed for real-time pose estimation and remain reliable under moderate lighting variation, partial occlusion, and image noise. Their square, high-contrast design supports accurate six-degree-of-freedom (6-DoF) pose estimation with minimal computational overhead \cite{garrido2014aruco}, and recent advances such as DeepArUco++ further improve robustness in low-contrast settings \cite{berral2024deeparuco}. Importantly, the use of uniquely identifiable markers enables \textit{selective beam redirection}, ensuring that only authenticated devices are tracked and served. This adds an additional level of user specificity while mitigating unintended signal leakage.

\begin{figure}[t]
    \centering
    \includegraphics[width=0.8\linewidth]{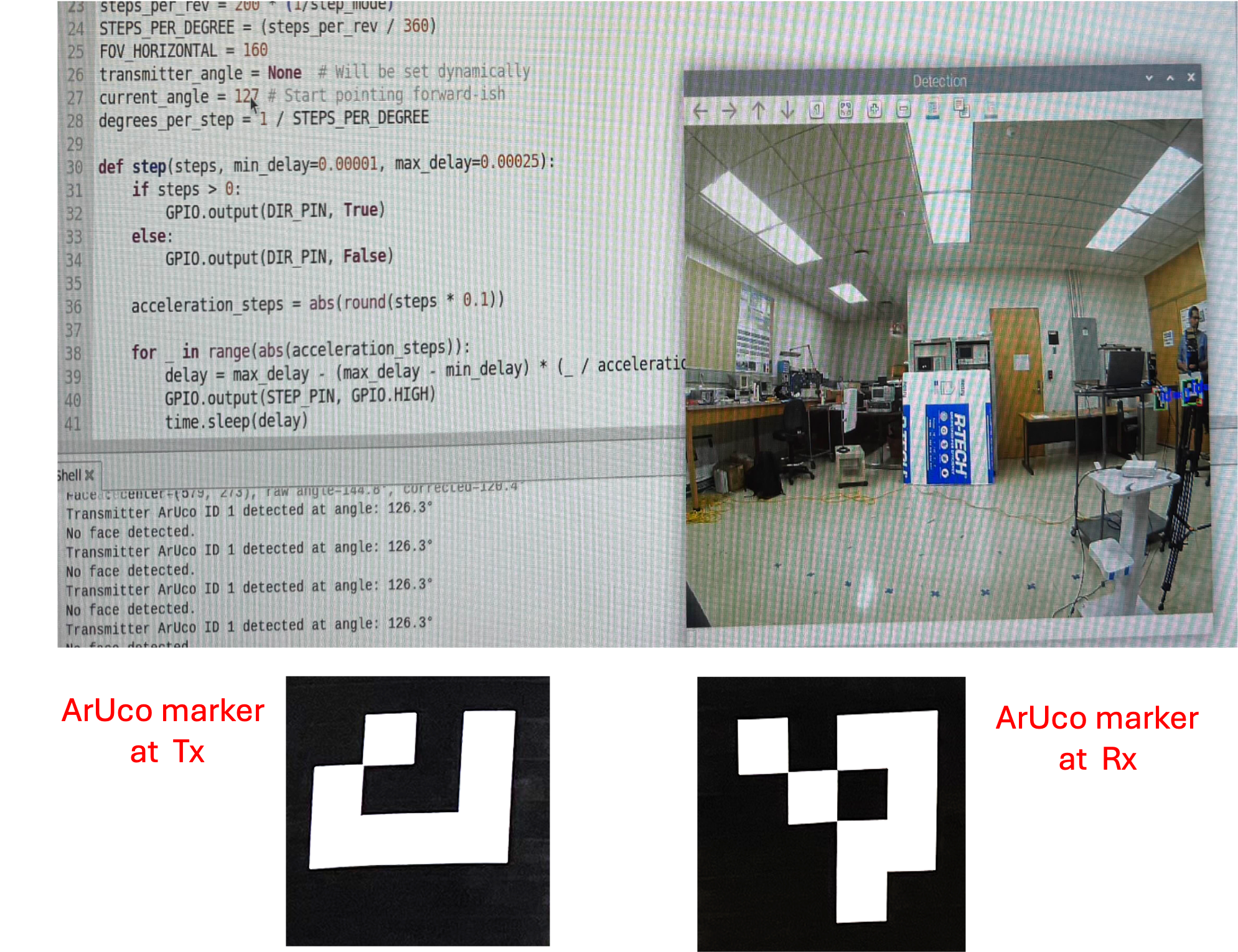}
    \caption{Camera view from the reflector system. ArUco markers on transmitter and receiver are detected (far right), enabling real-time orientation for optimal coverage.}
    \label{fig:camimage}
\end{figure}

\section{System Model}

We consider a millimeter-wave communication system in which a \textit{passive metallic reflector} is deployed to enhance connectivity between a transmitter and a receiver, particularly in NLoS scenarios. The transmitter and receiver are tagged with known ArUco markers, and the direct LoS path is assumed to be blocked, as illustrated in Fig.~\ref{fig:systemmodel}. In this setup, the reflector functions as an intelligent passive relay, dynamically redirecting the signal from transmitter to receiver through controlled alignment.

The reflector consists of a flat, perfectly conducting plate mounted on a stepper motor and equipped with a monocular camera. When both markers fall within the camera’s field of view, the reflector actively adjusts its orientation to maximize specular redirection. If a marker is not detected, the reflector defaults to static mode, behaving as a conventional passive surface. Although designed primarily for NLoS operation, the system can also enhance LoS links by improving robustness or enabling constructive multipath.

\subsection{Power Model with Radar Cross Section}

Under far-field and specular reflection conditions, the received power $P_{\text{Rx}}$ at the receiver is given by the bistatic radar equation~\cite{Rappaport2002,radarcrossection}
\begin{equation}
P_{\text{Rx}} = \frac{P_{\text{Tx}} G_{\text{Tx}} G_{\text{Rx}} \lambda^2 \sigma(\theta_i, \theta_r)}{(4\pi)^3 d_1^2 d_2^2},
\end{equation}
where $P_{\text{Tx}}$ is the transmit power, $G_{\text{Tx}}$ and $G_{\text{Rx}}$ are the antenna gains, $\lambda$ is the carrier wavelength, and $d_1$ and $d_2$ are the distances from transmitter to reflector and from reflector to receiver, respectively. The term $\sigma(\theta_i, \theta_r)$ denotes the bistatic radar cross section (RCS), which depends on the incident angle $\theta_i$ of the incoming wave relative to the reflector normal and the reflection angle $\theta_r$ of the reradiated wave.  

For a perfectly conducting flat plate of area $A$, the RCS under specular reflection is approximated as:
\begin{equation}
\sigma(\theta_i = \theta_r) \approx \frac{4\pi A^2 }{\lambda^2} \cos\!\left(\frac{\beta}{2}\right),
\end{equation}
where $\beta$ is the bistatic angle subtended at the reflector. The received power is therefore highly sensitive to reflector orientation, with maximum power achieved when $\beta \to 0$, i.e., when transmitter and receiver are symmetrically aligned with respect to the reflector normal.

\subsection{Reflection Control Strategy}

To maximize received power, the reflector minimizes $\beta$ by aligning such that $\theta_i \approx \theta_r \approx \beta/2$. The onboard camera detects ArUco markers, estimates their angles of arrival, and commands the stepper motor to reorient accordingly. This adaptive control strategy enables robust redirection of mmWave signals in complex indoor or obstructed environments.

\begin{algorithm}[t]
\caption{Vision-Based Reflector Orientation via Dual ArUco Marker AoA Estimation}
\label{alg:reflector_orientation}
\begin{algorithmic}[1]
\Require Image width $W$, FoV $\Theta_{\text{FOV}}$, markers $M_T$ (Tx), $M_R$ (Rx), current orientation $\theta_o$, microstepping $m$, steps/rev $N_{\text{step}}$
\Ensure Corrected AoAs $(\theta_T, \theta_R)$ and updated reflector orientation
\State Capture RGB frame $f_k$
\State Detect $\{M_T, M_R\}$ with \texttt{cv2.aruco.detectMarkers()}
\For{each marker $M_i \in \{M_T, M_R\}$}
    \State Compute pixel centroid $x_i$
    \State Estimate raw AoA: $\theta_{\text{raw}} \gets \tfrac{x_i}{W} \cdot \Theta_{\text{FOV}}$
    \State Apply distortion correction via Eq.~(\ref{eq:distortr}) to obtain $\theta_i$
\EndFor
\State Assign $\theta_T \gets \theta_{M_T}$, $\theta_R \gets \theta_{M_R}$
\State Compute $\theta_{\text{ref}} \gets \tfrac{\theta_T + \theta_R}{2}$ \Comment{bisect angles}
\State Compute step size $\Delta\theta \gets 360/(N_{\text{step}} \cdot m)$
\State Compute required steps $S \gets \operatorname{round}\!\big((\theta_{\text{ref}} - \theta_o)/\Delta\theta\big)$
\State Command motor with $|S|$ steps
\State Update $\theta_o \gets \theta_{\text{ref}}$
\State \Return $(\theta_T, \theta_R)$
\end{algorithmic}
\end{algorithm}

\section{Vision-Based Reflector Orientation Using ArUco Markers}

This section presents the vision-based pipeline for orienting a passive reflector using fiducial markers. A monocular camera captures image frames and identifies ArUco markers attached to the transmitter and receiver. Their pixel locations are mapped to angular bearings, and the reflector is reoriented so that its surface normal bisects the incident and reflected paths, minimizing the bistatic angle $\beta$ and maximizing specular reflection.

\subsection{AoA Estimation via Pixel-to-Angle Conversion}
Let $x$ denote the horizontal pixel coordinate of a detected marker centroid and $W$ the camera frame width. For a camera with horizontal field of view $\Theta_{\text{FOV}}$, the raw bearing angle is
\begin{equation}
\theta_{\text{raw}} = \frac{x}{W}\,\Theta_{\text{FOV}}.
\end{equation}
This linear model assumes a direct mapping between pixel position and angle. To account for radial distortion in wide-angle or low-cost lenses, we apply the nonlinear correction as follows
\begin{equation}
\theta = 43.5 \cdot \tan^{-1}\!\left(\frac{\theta_{\text{raw}} - \theta_c}{4000}\right) \cdot \frac{180}{\pi} + \theta_c, \label{eq:distortr}
\end{equation}
where $\theta_c = \Theta_{\text{FOV}}/2$ is the optical center. This empirically tuned transformation approximates the S-shaped distortion curve typical of barrel distortion. While less precise than calibration-based methods, it offers a lightweight alternative suitable for real-time embedded systems.

\subsection{Angle-to-Step Conversion}
The reflector is actuated by a stepper motor with $N_{\text{step}} = 200$ full steps per revolution and $m$ micro-steps. The angular resolution is
\begin{equation}
\Delta\theta = \frac{360}{N_{\text{step}}\,m}.
\end{equation}
Given the current orientation $\theta_o$ and target orientation $\theta_{\text{ref}}$, the motor executes
\begin{equation}
S = \operatorname{round}\!\left( \frac{\theta_{\text{ref}} - \theta_o}{\Delta\theta} \right)
\end{equation}
steps to align the reflector.

\subsection{Reflector Orientation Calculation}
Following Snell’s law, the reflector normal is set to bisect the incoming and outgoing rays. The desired orientation is
\begin{equation}
\theta_{\text{ref}} = \frac{\theta_{T} + \theta_{R}}{2}, \label{eq:snell}
\end{equation}
where $\theta_T$ and $\theta_R$ are the AoAs of the transmitter and receiver, respectively.

\subsection{Stepper Motor Calibration}
Since the stepper motor does not provide absolute position feedback, an initial calibration is required. The reflector is manually aligned to a reference orientation (e.g., $0^\circ$), the motor coupling is re-engaged, and $\theta_o$ is initialized accordingly. This one-time setup takes less than five seconds and enables repeatable operation.

The complete procedure, including marker detection, angle estimation with distortion correction, reflector orientation calculation, and stepper motor actuation, is summarized in Algorithm~\ref{alg:reflector_orientation}, which provides a lightweight and repeatable pipeline for real-time reflector alignment in embedded systems.

\begin{figure}[t]
    \centering
    \includegraphics[width=0.8\linewidth]{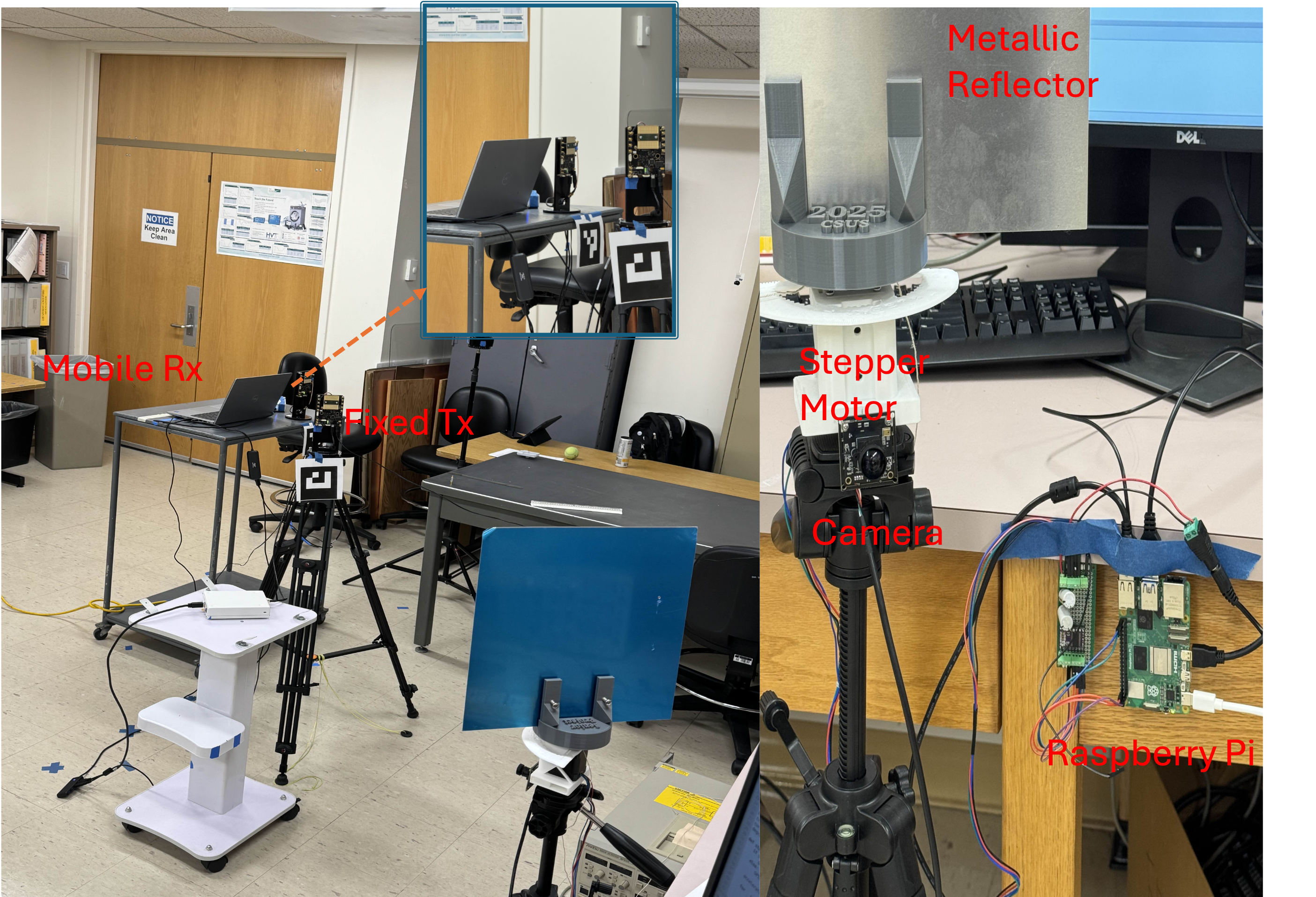}
    \caption{Prototype architecture. A Raspberry Pi captures camera input, detects ArUco markers, computes the reflector orientation, and commands the stepper motor for alignment.}
    \label{fig:setup}
\end{figure}

\begin{table*}[t]
\centering
\caption{Measured versus vision based AoA estimates at 2\,m, 3\,m, and 4\,m. Results show sub degree accuracy near boresight and errors of only $\pm 3$–$6^\circ$ at wider angles, which remain within typical mmWave half power beamwidths. Missing entries (--) indicate angles not tested at a given distance.}
\label{tab:aoa_all}
\begin{tabular}{c|cc|cc|cc}
\hline
Measured ($^\circ$) & \multicolumn{2}{c|}{2 m} & \multicolumn{2}{c|}{3 m} & \multicolumn{2}{c}{4 m} \\
 & Estimated ($^\circ$) & Error ($^\circ$) & Estimated ($^\circ$) & Error ($^\circ$) & Estimated ($^\circ$) & Error ($^\circ$) \\
\hline
-35 & -38.7 & -3.7 &   --   &   --   &   --   &   --   \\
-30 & -34.4 & -4.4 & -32.7 & -2.7 &   --   &   --   \\
-25 & -29.3 & -4.3 & -27.5 & -2.5 & -27.6 & -2.6 \\
-20 & -22.6 & -2.6 & -22.9 & -2.9 & -22.1 & -2.1 \\
-15 & -16.0 & -1.0 & -18.7 & -3.7 & -16.0 & -1.0 \\
-10 & -11.0 & -1.0 & -12.6 & -2.6 & -11.0 & -1.0 \\
-5  & -5.9  & -0.9 & -6.2  & -1.2 & -6.5  & -1.5 \\
0   & 0.5   & 0.5  & -0.1  & -0.1 & -0.4  & -0.4 \\
5   & 6.6   & 1.6  & 5.8   & 0.8  & 7.5   & 2.5 \\
10  & 13.3  & 3.3  & 10.5  & 0.5  & 10.8  & 0.8 \\
15  & 17.7  & 2.7  & 16.1  & 1.1  & 16.2  & 1.2 \\
20  & 22.9  & 2.9  & 22.0  & 2.0  & 21.8  & 1.8 \\
25  & 29.9  & 4.9  & 28.9  & 3.9  & 27.9  & 2.9 \\
30  & 33.2  & 3.2  & 33.7  & 3.7  &   --   &   --   \\
35  & 40.7  & 5.7  & 38.4  & 3.4  &   --   &   --   \\
40  & 44.9  & 4.9  &   --   &   --   &   --   &   --   \\
45  & 49.0  & 4.0  &   --   &   --   &   --   &   --   \\
\hline
\end{tabular}
\end{table*}

\subsection{AoA Estimation Error Analysis}

To evaluate the accuracy of the proposed vision based AoA estimation, we conducted controlled measurements at distances of 2\,m, 3\,m, and 4\,m from the camera. A subject holding an ArUco marker was positioned at boresight ($0^\circ$) and walked laterally in parallel to the camera axis. At each position, the angular displacement was measured with a protractor and compared against the estimated AoA from the vision pipeline. The process continued until the marker was either outside the field of view or unreadable.

Table~\ref{tab:aoa_all} summarizes the measured and estimated AoAs at three distances. At 4\,m, the estimation error remained within $\pm 3^\circ$, with sub-degree accuracy near boresight. At 3\,m, the error increased slightly at wider angles, reaching 3.9$^\circ$ at $+25^\circ$. At 2\,m, errors became more pronounced, particularly at large off-axis angles where distortion effects dominate; the maximum error was approximately 5.7$^\circ$ at $+35^\circ$. Across all cases, however, the error stayed below $1^\circ$ near the optical axis, confirming robustness in the most critical region for reflector alignment.  Two key trends are evident (i) \textit{distance dependence}: at shorter distances, the same angular displacement corresponds to larger pixel shifts, amplifying distortion and increasing error at wide angles; and (ii) \textit{angle dependence}: errors remain minimal near the center but grow at off-axis positions due to lens distortion and reduced marker resolution toward the image edges, even after distortion compensation. 

\begin{figure}[t]
    \centering
    \includegraphics[width=0.8\linewidth]{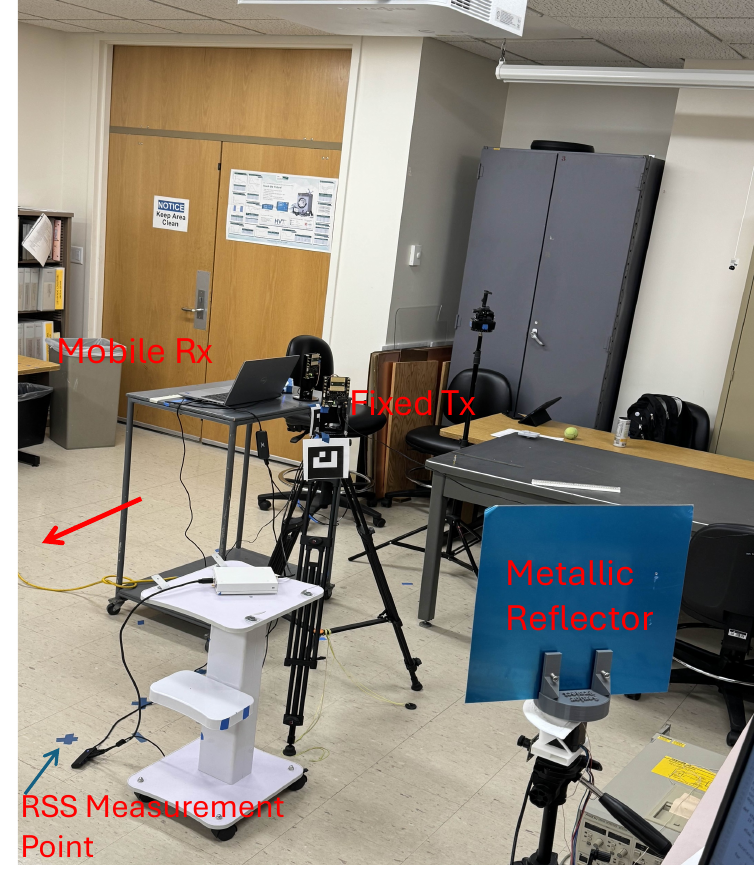}
   \caption{Measurement  and environment setup illustrating the fixed transmitter, the initial position of the mobile receiver,  and the passive reflector. }
    \label{fig:setup0}
\end{figure}

\section{Prototype Architecture and \\ Experimental Setup}
\label{sec:prototype}

The architecture of the proposed vision-guided reflector prototype is shown in Fig.~\ref{fig:setup}. The prototype employs a $300\,\mathrm{mm} \times 300\,\mathrm{mm}$ aluminum flat-plate reflector mounted on a two-axis Dynamixel gimbal. Actuation is provided by a 0.9$^\circ$ stepper motor operated in $1/16$ micro-stepping mode, achieving an angular resolution of $\Delta\theta = 0.1125^{\circ}$. The motor is driven by a DRV8824 breakout board, with control and vision processing executed on a Raspberry Pi~4 running Ubuntu~22.04 and OpenCV~4.10.0. Visual feedback is obtained from an Arducam OV9281 global-shutter camera ($640\times480$ resolution, $120^{\circ}$ horizontal FoV), which continuously detects ArUco markers affixed to the transmitter and receiver. The Raspberry Pi performs both pose estimation and reflector steering in real time.   The mmWave testbed consists of two 60\,GHz Sivers phased-array modules integrated with NI-USRP baseband hardware. The transmitter (tagged with ArUco ID~1) is fixed and configured for directional transmission toward the reflector, while the receiver (tagged with ArUco ID~0) operates in omnidirectional mode.

Experiments were conducted in an indoor laboratory under NLoS conditions, as illustrated in Fig.~\ref{fig:setup0}. The transmitter and receiver were initially positioned 5\,ft and 6\,ft from the reflector, respectively, each forming a $45^\circ$ angle with the reflector surface. The receiver was then moved laterally in 0.5\,ft increments up to 9.5\,ft, following the path indicated in Fig.~\ref{fig:setup0}. The relative received power was recorded at each marked position.  To evaluate performance, four configurations were considered: a LoS baseline with directional transmission and an omnidirectional receiver, an NLoS baseline with no reflector, an NLoS case with a manually aligned static reflector, and an NLoS case with the proposed vision-guided reflector.  The primary performance metric is the communication outage probability, defined as the probability that the received relative power $\gamma$ falls below a threshold $\gamma_{\text{th}}$, i.e.,  $   P_o = \mathbb{P}(\gamma \leq \gamma_{\text{th}}).$ The objective of the proposed system is to minimize $P_o$ by dynamically steering the reflector to maintain strong NLoS links.

\begin{figure}[t]
\hspace{-7mm}
  \centering
  \includegraphics[width=0.9\linewidth]{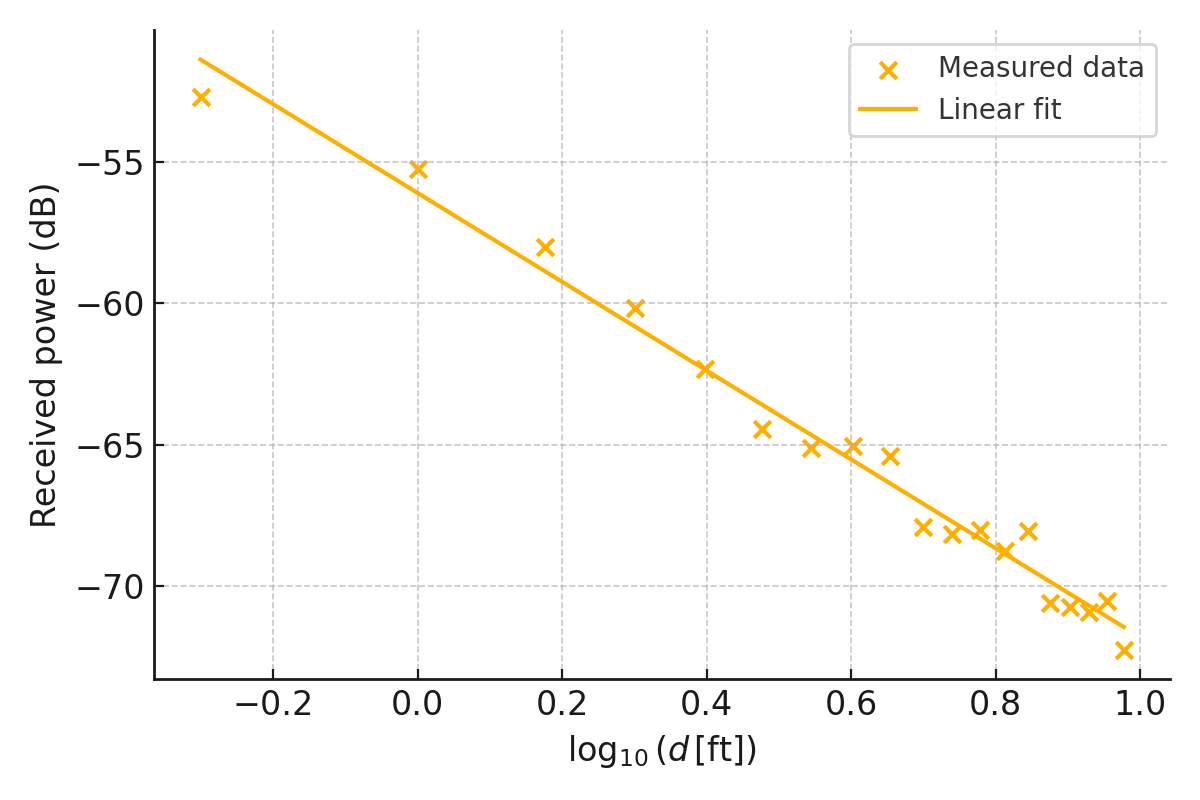}
  \caption{Received (relative) power versus $\log_{10}(d)$ for the line of sight baseline (directional TX, omni directional RX).}
  \label{fig:los_logfit}
\end{figure}

\section{Experimental Results and Discussion}
\label{sec:results}

This section presents the experimental evaluation of the proposed vision-guided reflector system against three baselines: (i) LoS link coverage, (ii) NLoS with no reflector, and (iii) a manually aligned reflector. 

In Fig.~\ref{fig:los_logfit} we plot the received relative power LoS baseline with a directional TX and omnidirectional RX, plotted against $\log_{10}(d)$,  where $d$ is the TX-RX distance in feet (see Fig. \ref{fig:setup0}). The regression $P(d)=a+b\log_{10}(d)$ yields $a=-56.10$\,dB and $b=-15.71$\,dB/decade, corresponding to a path loss exponent of $n=1.57$ with $R^2=0.981$.  This result aligns with prior indoor mmWave LoS studies ($n\approx 1.3$--2.0)~\cite{MacCartney2015IndoorOffice,Sun2015VTCIndoorOffice}, validating the measurement methodology and establishing a baseline reference for reflector-assisted NLoS performance.

\begin{figure}[t]
\hspace{-5mm}
  \centering
  \includegraphics[width=1\linewidth]{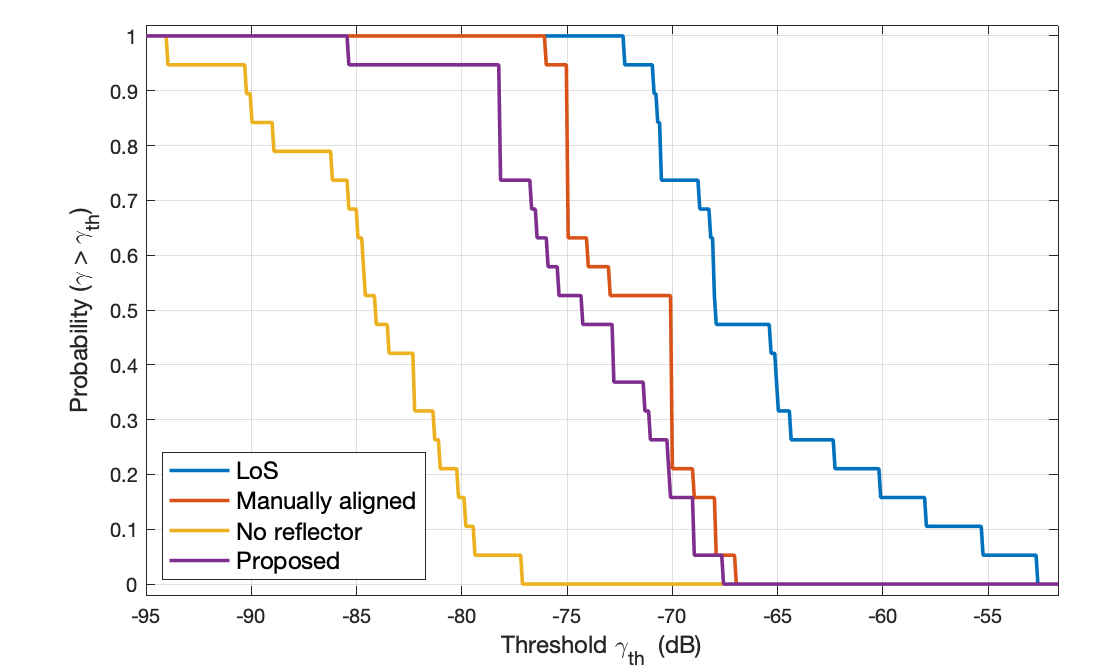}
\caption{Complementary CDF of the received relative power for four cases over all distances: line of sight,  manually aimed reflector, no reflector, and the proposed vision guided reflector. The curve at each threshold shows $\mathbb{P}(\gamma > \gamma_{\mathrm{th}})$.}
  \label{fig:ccdf_compare}
\end{figure}

The complementary CDF of received power across all distances highlights clear performance differences among the four configurations as shown in Fig. \ref{fig:ccdf_compare}. The LoS case offers the strongest performance; for example, at $-75$\,dB relative power,  it maintained 100\% link availability. The manually aligned reflector follows with $\approx$63\% availability, while the proposed vision-guided reflector achieves $\approx$53\%. By contrast, the no-reflector case provides essentially no availability at the same threshold. Fluctuations in received power are expected in indoor environments, and this trend is consistent with vision-aided RIS results reported in \cite{ouyang2023cvris}. Overall, these results demonstrate that vision-guided steering recovers most of the reflector gain without requiring manual alignment, thereby enabling robust coverage in NLoS conditions where the baseline link would otherwise fail.

\begin{figure}[t]
\hspace{-3mm}
    \centering
    \subfloat[]{%
        \includegraphics[width=0.48\linewidth]{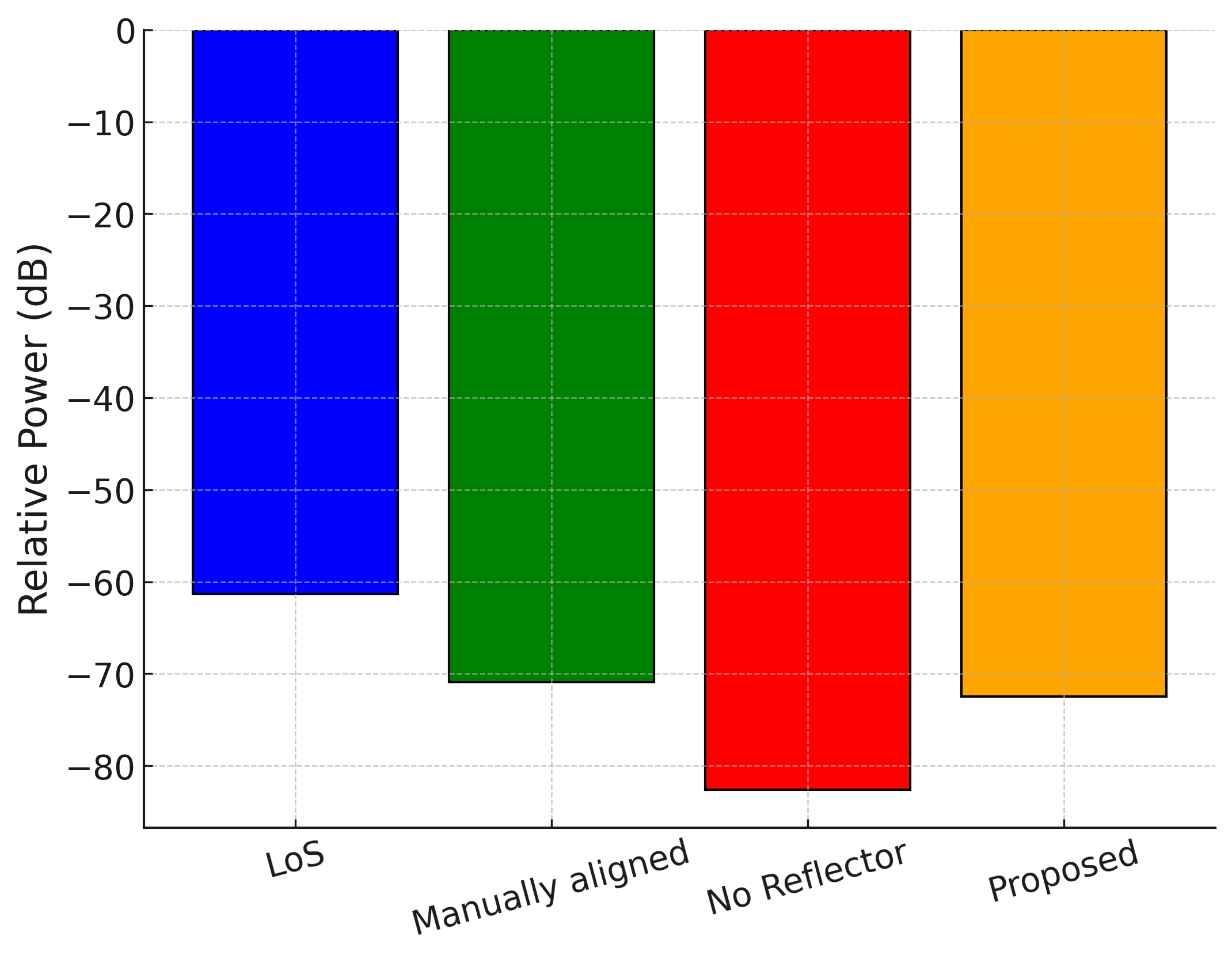}%
        \label{fig:avg_power}
    }
    \hfill
    \subfloat[]{%
        \includegraphics[width=0.48\linewidth]{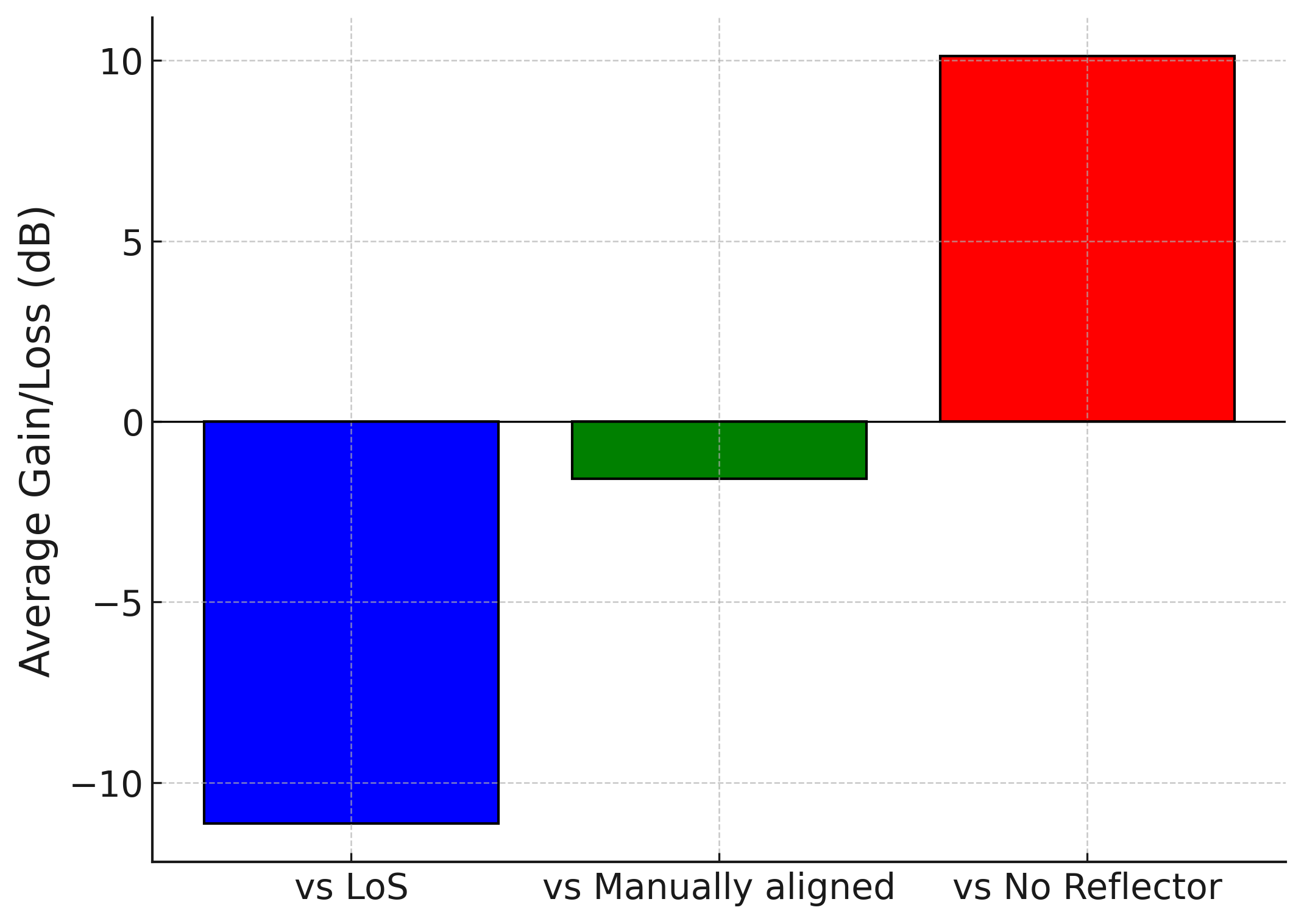}%
        \label{fig:avg_gain}
    }
    \caption{Performance comparison: 
    (a) Average received relative power across all configurations; 
    (b) Average gain/loss of the proposed approach relative to LoS, manually aligned reflector, and no-reflector baseline.}
    \label{fig:avg_results}
\end{figure}

The averaged metrics highlight clear differences across configurations, as shown in Fig.~\ref{fig:avg_results}. Subfigure~\ref{fig:avg_power} shows the mean received power, underscoring the large gap between LoS and NLoS cases. The proposed reflector recovers much of this loss and closely approaches the manually aligned reflector. Subfigure~\ref{fig:avg_gain} presents the average gain/loss relative to the baselines where the proposed vision-guided reflector delivers an average $\approx$10\,dB improvement over the no-reflector case, remains within 1--3\,dB of the manually aligned reflector, yet lags $\approx$9--11\,dB behind LoS.

The modest shortfall relative to manual alignment can be attributed to several small but compounding sources of error, including inaccuracies in marker-based pose estimation, initial calibration offsets, the angular estimation errors reported in Table~\ref{tab:aoa_all}, and the limited camera field of view, all of which reduce steering precision. In contrast to manual alignment, which fixes the reflector orientation once under controlled conditions, our system must continuously infer geometry from visual cues, making it more susceptible to sensor noise, occlusion, and distortions at wider angles. Moreover, unlike the LoS baseline, which benefits from direct propagation with minimal reflection loss, any reflector-based approach inevitably incurs additional path loss and is therefore fundamentally limited in absolute received power.  Overall, although the proposed vision guided system cannot replicate the ideal conditions of LoS propagation or the precision of manual alignment, it still provides robust NLoS coverage with minimal infrastructure and control overhead, underscoring a practical balance between performance and autonomy.


\section{Conclusion and Future Directions}
\label{sec:conclusion}

This work presented a vision-guided passive reflector prototype for enhancing mmWave links in dynamic NLoS environments. By detecting fiducial markers and autonomously steering a gimbal-mounted reflector, the system achieves real-time alignment without RF feedback or additional infrastructure. Indoor 60\,GHz experiments demonstrated received relative power improvements of up to 17\,dB (with an average gain of $\approx$10\,dB) compared to the no-reflector baseline, and an increase in link availability from essentially 0\% to over 50\% at a $-75$\,dB threshold. These results establish the proposed approach as a low-cost, energy-efficient alternative to RIS panels or relays for short-range coverage (e.g., Wi-Fi, AR/VR, IoT).

Several practical constraints remain, including limited camera field-of-view, marker visibility under occlusion or poor lighting, processing latency, spoofing risks, and single-user operation. The use of an omnidirectional RX antenna also provides a conservative lower bound, since joint beam-tracking with directional RX antennas was not evaluated. Pose estimation errors near the field-of-view edges further reduce accuracy.

Future work will explore higher-resolution or markerless vision methods, wide-angle cameras, joint reflector--receiver beam coordination, and hardware acceleration to reduce actuation latency. Addressing multi-user scalability and secure fiducial encoding are key challenges, while drone-mounted or distributed reflectors represent promising options for rapid deployment. Overall, vision-guided reflectors offer a practical, low-cost means to improve mmWave coverage, with strong potential for future system-level co-design.

\section*{Acknowledgment}
This material is based on work supported by the National Science Foundation under grant No. NSF-2243089.

\balance

\end{document}